\documentclass[journal]{IEEEtran}
\usepackage[section]{placeins}
\usepackage{graphicx,subfigure}
\usepackage{amsmath}
\usepackage{amsfonts}
\usepackage{caption2}
\usepackage{amssymb}
\usepackage{varwidth}
\usepackage[rightbars,color]{changebar}
\usepackage{epsfig}
\usepackage{multirow}
\usepackage{booktabs}
\usepackage[linesnumbered,ruled,vlined]{algorithm2e}
\usepackage{array}
\usepackage{rotating}

\usepackage{epsfig,amsfonts,multirow}

\ifCLASSINFOpdf

\else

\fi

\begin{document}

\title{Adaptive Deep Pyramid Matching for Remote Sensing Scene Classification}
\author{~Qingshan~Liu,~\IEEEmembership{Senior~Member,~IEEE}, ~Renlong~Hang, ~Huihui~Song, ~Fuping~Zhu,\\ ~Javier~Plaza, ~\IEEEmembership{Senior~Member,~IEEE} and ~Antonio~Plaza, ~\IEEEmembership{Fellow,~IEEE}
\thanks{Q. Liu, R. Hang and H. Song are with the Jiangsu Key Laboratory of Big Data Analysis Technology, Jiangsu Collaborative Innovation Center on Atmospheric Environment and Equipment Technology,
Nanjing University of Information Science and Technology, Nanjing 210044, China (e-mail: qsliu$@$nuist.edu.cn, renlong\_hang$@$163.com).

F. Zhu is with Beijing Electro-Mechanical Engineering Institute, Beijing 100083, China.

J. Plaza and A. Plaza are with the Hyperspectral Computing Laboratory, Department of Computer Technology and Communications, University of Extremadura, Caceres E-10071, Spain.
}}%
\markboth{Submitted to IEEE Transactions on Geoscience and Remote Sensing}%
{Shell \MakeLowercase{\textit{et al.}}:}

\maketitle

\begin{abstract}
Convolutional neural networks (CNNs) have attracted increasing attention in the remote sensing community. Most CNNs only take the last fully-connected layers as features for the classification of remotely sensed images, discarding the other convolutional layer features which may also be helpful for classification purposes. In this paper, we propose a new adaptive deep pyramid matching (ADPM) model that takes advantage of the features from all of the convolutional layers for remote sensing image classification. To this end, the optimal fusing weights for different convolutional layers are learned from the data itself. In remotely sensed scenes, the objects of interest exhibit different scales in distinct scenes, and even a single scene may contain objects with different sizes. To address this issue, we select the CNN with spatial pyramid pooling (SPP-net) as the basic deep network, and further construct a multi-scale ADPM model to learn complementary information from multi-scale images. Our experiments have been conducted using two widely used remote sensing image databases, and the results show that the proposed method significantly improves the performance when compared to other state-of-the-art methods.

\end{abstract}
\IEEEpeerreviewmaketitle

\section{Introduction}
With the development of remote sensing technology, large amounts of Earth-observation images with high resolution can be acquired nowadays from satellites and airplanes. Automatic and accurate interpretation of such massive image repositories has become an urgent demand. Among various applications, including land-resource management and urban planing, remote sensing scene classification is a fundamental task and has been an active research topic in the past decades. Different from other images, remotely sensed scenes exhibit some particular characteristics. For instance, there are various types of objects with different sizes, colors and orientations in a single scene. The object of interest is often surrounded by other materials. Besides, the increasing spatial resolution makes the detail information more clear, thus increasing the inter-class similarities and intra-class variances. Therefore, it is difficult to learn robust and discriminative representations from remotely sensed scenes.

In the past few years, a large amount of feature representation models have been proposed for scene classification. One of the most popularly used model is the bag of visual words (BoVW) \cite{sivic2003, csurka2004, fei2005}, which generally includes the following three steps: 1) extracting low-level visual features, such as scale-invariant feature transform (SIFT) \cite{lowe2004} and histogram of oriented gradients (HOG) \cite{dalal2005} descriptors; 2) clustering features to form visual words (clustering centers) by using $k$-means or other methods; 3) mapping visual features to the closest word and generating a mid-level feature representation by word histograms. This model and its variants have been extensively investigated in the field of remote sensing  \cite{yang2008, lienou2010, cheriyadat2014}.

Despite the effectiveness of the BoVW model, it provides an orderless collection of local descriptors (regardless of the spatial information). To overcome this drawback, the spatial pyramid matching (SPM) model was proposed in \cite{lazebnik2006}. This model first partitions the original image into different levels of resolutions. Secondly, for each level of resolution, the histograms of local features are extracted inside each spatial bin. Lastly, the spatial histograms are weighted via the pyramid matching kernel \cite{grauman2005}.
Since an absolute reference frame usually does not exist for remote sensing images, the relative spatial arrangement of image elements becomes very important.
Accordingly, the spatial pyramid co-occurrence model was proposed to characterize both the photometric and geometric information of an image in \cite{yang2011}. Unlike the division of the image into uniform cells in \cite{lazebnik2006} and \cite{yang2011}, the authors in \cite{jiang2012} proposed a randomized spatial partition to characterize various image layouts.

All the aforementioned methods are based on hand-crafted features, which heavily depend on the experience and domain knowledge of experts. Moreover, such features are difficult to exploit to create an optimal balance between discriminability and robustness. This is mainly due to the lack of considerations for the details of real data \cite{chen2014}. Deep learning (DL) algorithms \cite{lecun2015, bengio2013}, especially convolutional neural networks (CNNs) \cite{krizhevsky2012}, have shown tremendous potential to address this issue, as they can automatically learn the high-level semantic features from raw images in a hierarchical manner, which has drawn increasing attention in the remote sensing community \cite{luus2015, 7486259, zhao2016, romero2016}.

However, it is difficult to directly apply the CNNs to remote sensing scene classification problems, because there often exist millions of parameters to train, while the number of available training samples is small. The works in \cite{oquab2014, donahue2014, sharif2014, yosinski2014}
have indicated that the features extracted from the CNNs can be considered as generic descriptors.
Therefore, the image representations learned from CNNs on large-scale annotated datasets, such as ImageNet \cite{deng2009}, can be effectively transferred to a wide range of visual recognition tasks with a limited amount of training data. Motivated by this idea, the works in \cite{castelluccio2015, penatti2015, marmanis2016} validated the feasibility of employing the ImageNet pre-trained CNNs to remotely sensed scene classification. In \cite{castelluccio2015}, the authors employed the pre-trained CNNs and fine-tuned them on the scene datasets, showing impressive classification performance. In \cite{penatti2015}, the authors evaluated the generalization power of the CNN features from fully-connected layers in remote sensing image classification and showed state-of-the-art results on a public remotely sensed scene dataset.

Although transfer learning is able to alleviate the overfitting problem, there still exist some issues for CNNs-based remote sensing scene classification. The first one is that most of the CNNs only take the last fully-connected layers as features for subsequent classifications. It is not reasonable to directly discard the earlier convolutional layer features, which may be also helpful for classification purposes. Actually, features from convolutional layers are more generic than those from fully-connected layers \cite{yosinski2014}, and thus these features may be more suitable for transfer learning. In addition, convolutional layer features contain more spatial information, which offers advantages for image classification \cite{lazebnik2006, he2014} as compared to the activation of fully-connected layers. A few recent works have realized the importance of convolutional layer features \cite{chen2014, hu2015}, but they only use the last convolutional layer while ignoring other convolutional layers.

\begin{figure}
  \centering
  \includegraphics[scale = 0.7]{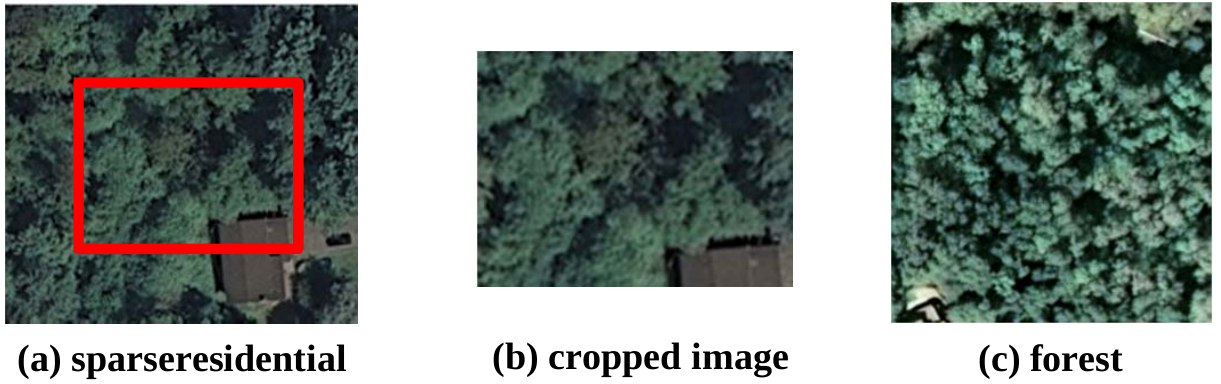}\\
  \caption{Loss of effective discriminative information by cropping an image to a fixed size. In (b), the cropped image of (a) just contains a small part of house, which loses the essential information for discriminating the \textit{sparseresidential} class from the \textit{forest} class in (c).}\label{ImageExample_cropping}
\end{figure}

\begin{figure}
  \centering
  \includegraphics[scale = 0.55]{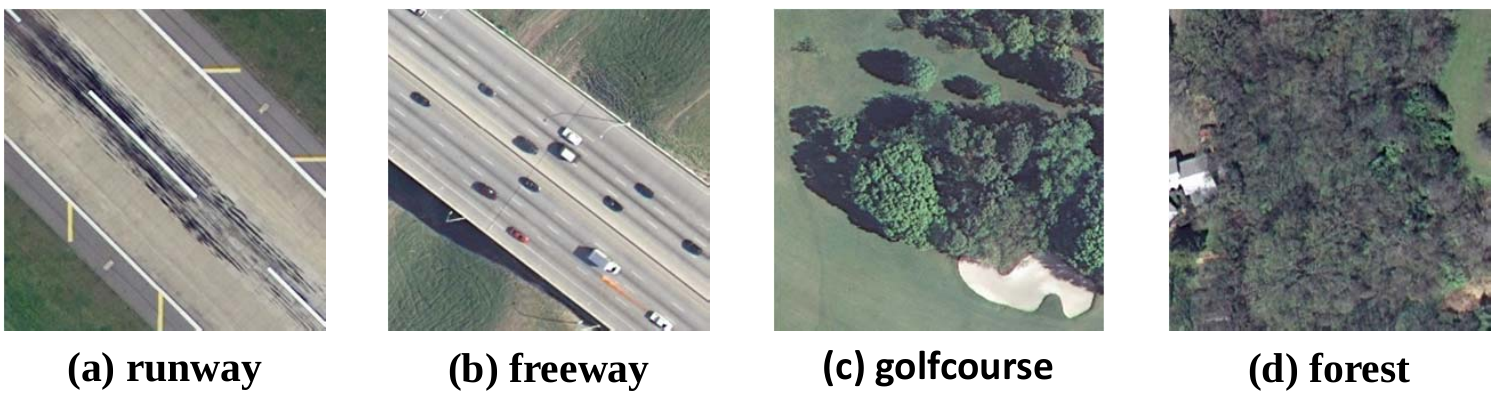}\\
  \caption{Loss of effective discriminative information by warping an image to a fixed size. The cars in (b) will disappear if warping the image to a small size, making it difficult to differentiate the \textit{runway} class in (a) and the \textit{freeway} class in (b). On the contrary, warping the image in (c) to a large size will enrich the information of trees, thus increasing the possibility to confuse the \textit{golfcourse} class in (c) as \textit{forest} class in (d).}\label{ImageExample_warping}
\end{figure}

Another issue is that the objects of interest generally have different scales in different remotely sensed scenes, and even a single scene may contain objects with different sizes. However, most of the prevalent CNNs require a fixed input image size (e.g., $227\times227$ pixels). One common solution is to warp or crop the original remotely sensed image to a pre-defined size, which inevitably causes the loss of effective discriminant information as illustrated in Fig.~\ref{ImageExample_cropping} (cropping example) and Fig.~\ref{ImageExample_warping} (warping example).

To address the aforementioned issues, in this paper we propose a new multi-scale Adaptive Deep Pyramid Matching (ADPM) model. Inspired by SPM, we consider features in all of the convolutional layers as a multi-resolution representation of the input image. Then, the pyramid matching kernel is used to combine them into a unified representation. Different from SPM, we replace the low-level descriptors as deep features, and the optimal fusing weights among different convolutional layers are learned from data itself, instead of pre-defined values.
Besides, to reduce the information loss caused by a fixed input image size, the multi-scale images are fed into the CNNs to learn complementary information from different scales.
Considering the computation cost of learning multi-scale deep features simultaneously, we select the CNN with Spatial Pyramid Pooling (SPP-net) as our basic deep network. SPP-net was proposed to address the size issue of input image \cite{he2014, he2015}. By adding an SPP layer before the first fully-connected layer, it allows input images with arbitrary sizes. Thus, one well-trained SPP-net can extract multi-scale deep features from multi-scale input images.

The remainder of this paper is organized as follows. In Section II, we present the proposed method in detail,
including the basic architecture of SPP-net and the feature fusing model ADPM. The experiments are reported in Section III, followed by the conclusions in Section IV.

\section{Methodology}
\begin{figure*}
  \centering
  \includegraphics[scale=0.7]{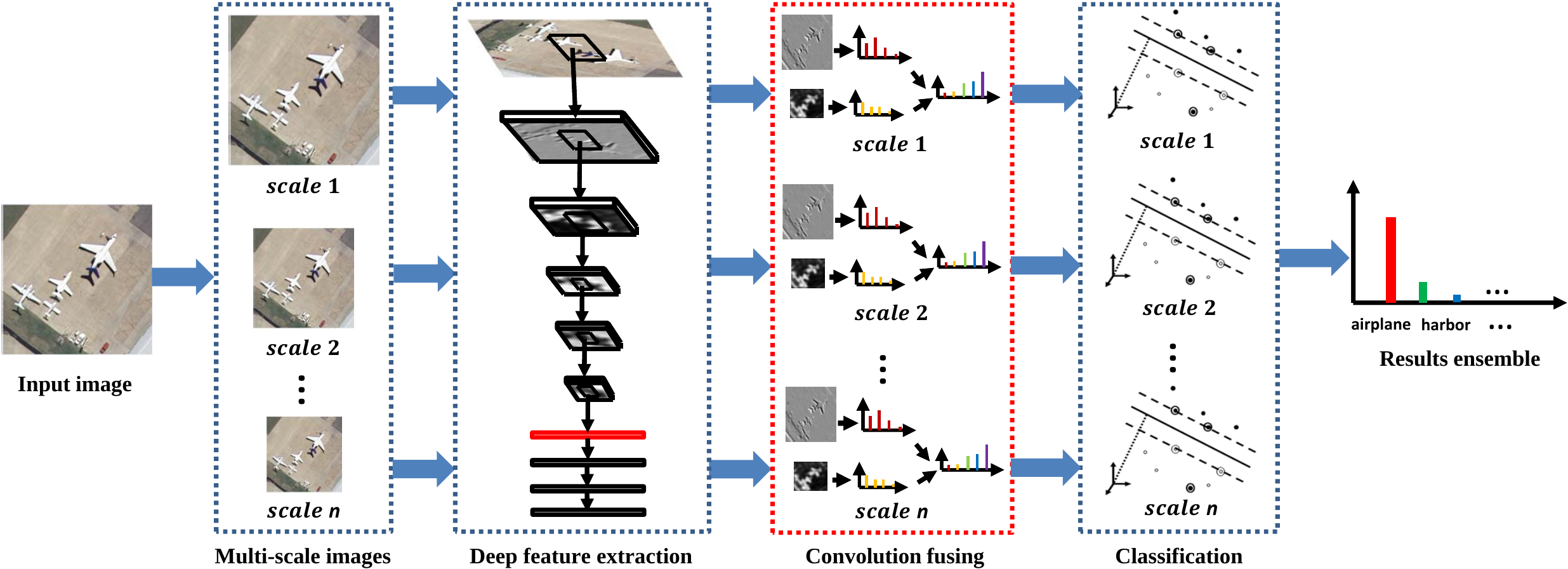}\\
  \caption{Flowchart of the proposed method.}\label{FlowChart}
\end{figure*}

The flowchart of the proposed method is shown in Fig.~\ref{FlowChart}, where the red box highlights the core ADPM algorithm. The whole procedure consists of four steps. First of all, an input image is warped into different scales. Secondly, the warped multi-scale images are fed into the SPP-net to extract  multi-scale deep features. Then, for each scale, the ADPM model is employed to fuse the extracted representations from all of the convolutional layers. Finally, the learned representation is fed into a support vector machine (SVM) to get a final classification result, and a majority voting strategy is used to integrate the multiple results from all scales. In the following subsections, we respectively present the multi-scale deep feature extraction method and the ADPM algorithm in detail.

\subsection{Multi-scale deep feature extraction}
\begin{figure*}
  \centering
  \includegraphics[scale = 0.7]{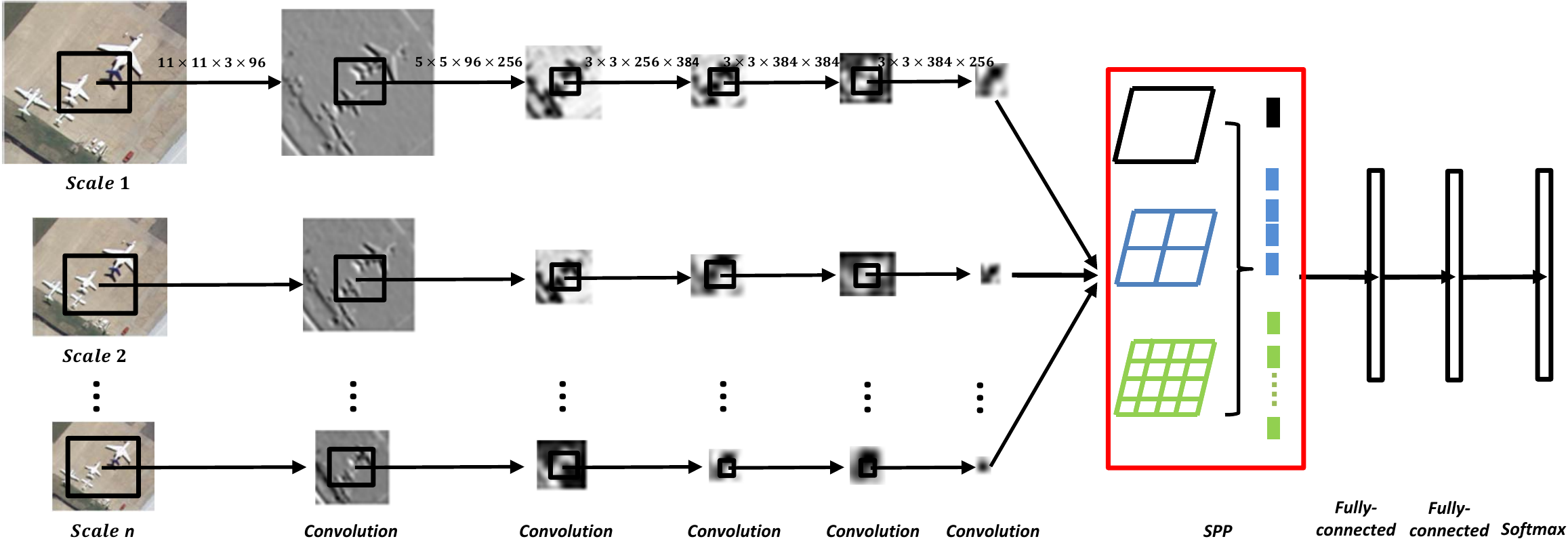}\\
  \caption{Architecture of the multi-scale deep feature extraction process.}\label{spp-net}
\end{figure*}

Since objects of interest often have different scales in remotely sensed scenes, we aim at learning multi-scale features to capture such property.
However, most of the CNNs require a fixed input image size. Therefore, it is difficult to extract multi-scale deep features simultaneously from one network. In this work, we propose to adopt a SPP-net based framework to address this issue.
The designed architecture is shown in Fig.$~$\ref{spp-net}, where the network contains five successive convolutional layers, an SPP layer highlighted in red color, two fully-connected layers, and a \textit{softmax} layer. The parameters between the two layers represent the kernel size (the first three numbers) and the number of kernels (the last number). Before the first fully-connected layer, SPP is used to pool the features from the last convolutional layer. Similar to SPM \cite{lazebnik2006}, we partition the feature maps into increasingly fine sub-regions, and pool the responses inside each sub-region via max pooling. Let us assume that the size of each feature map after the last convolutional layer is $a\times a$ pixels, and that each feature map is partitioned into $n\times n$ sub-regions. Then, SPP can be considered as convolution operator in a sliding-widow manner with window size $\textit{win} = \lceil a/n\rceil $ and stride $\textit{str} = \lfloor a/n\rfloor $ , where $\lceil \cdot\rceil$ and $\lfloor \cdot\rfloor $ denote ceiling and floor operators, respectively. Here, we choose a three-level SPP configuration by setting $n\times n$ as $1\times 1$, $2\times2$ and $4\times4$, respectively. The final output of SPP is to concatenate these three-level pooling results into a vector, resulting in a fixed length representation regardless of the input image size. Thus, input images with different sizes can share a single SPP-net.

To train an efficient and effective SPP-net, as suggested by \cite{oquab2014}, we firstly pre-train the network in \cite{krizhevsky2012} using the auxiliary ImageNet 2012 dataset (Source task), and then fine-tune the SPP-net by employing the training samples from scene images (Target task). The training procedure of the source task is carried out via the open source Caffe DCNN library \cite{jia2014}. After the pre-training of the source task, the weight parameters learned in the five convolutional layers are transferred to the target task and kept fixed. For the target task, we only need to fine-tune the last three layers (i.e., two fully-connected layers and one output layer).

\subsection{Adaptive deep pyramid matching (ADPM)}

\begin{figure}
  \centering
  \includegraphics[scale=0.6]{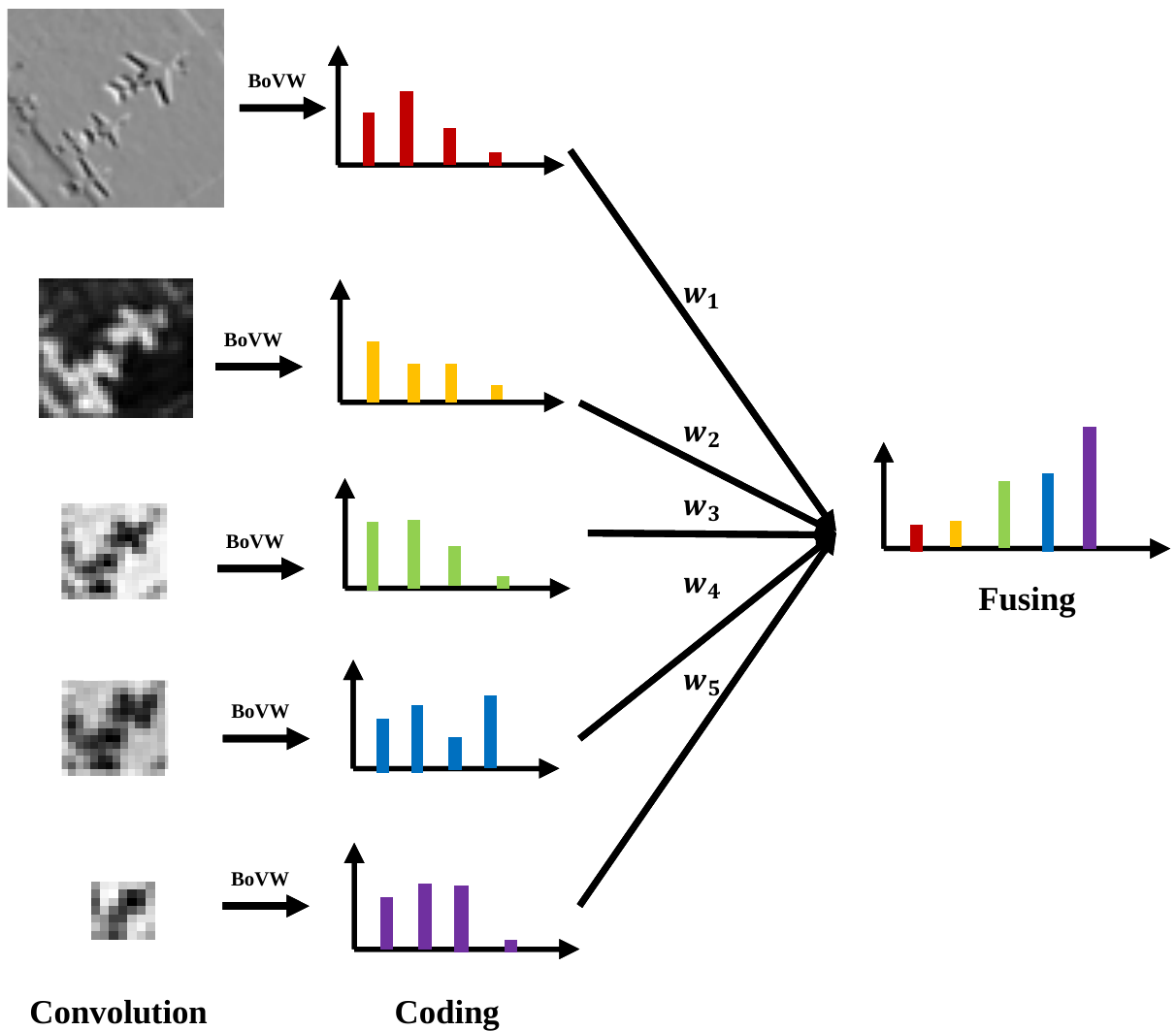}\\
  \caption{Flowchart of ADPM.}\label{ADPM}
\end{figure}

As demonstrated in Fig.~\ref{spp-net}, the convolutional layer features can be regarded as the multi-resolution deep representations of an input image. In this subsection, we introduce an ADPM model to optimally fuse these features. The flowchart of ADPM is shown in Fig.~\ref{ADPM}. Let us assume that a three-dimensional matrix $\mathbf{F}_{1,\ell}\in R^{n_{\ell}\times n_{\ell}\times p_{\ell}}$ denotes the $\ell-th$ layer feature map of image $\mathbf{I}_{1}$; then, at each location $(i,j), 1\leq i\leq n_{\ell}$ and $1\leq j\leq n_{\ell}$,
$\mathbf{f}_{1,\ell}^{(i,j)}$ defines a $p_{\ell}$-dimensional representation for a local patch of the image $\mathbf{I}_{1}$. In this way, we obtain $n_{\ell}\times n_{\ell}$ local feature vectors for image $\mathbf{I}_{1}$ at layer $\ell$. Similar to BoVW, we use $k$-means to cluster all of these features into $D$ centers $\mathbf{c}_{\ell}^{1}, \cdots, \mathbf{c}_{\ell}^{D}$ to form a vocabulary, and each $\mathbf{f}_{1,\ell}^{(i,j)}$ is assigned to its nearest visual word $\mathbf{c}_{\ell}^{d}, 1\leq d \leq D$. Then, $\mathbf{F}_{1,\ell}$ is converted to a histogram representation:
\begin{equation}\label{}
\mathbf{H}_{1,\ell} = [\sum_{i,j}\delta(\mathbf{f}_{1,\ell}^{(i,j)},\mathbf{c}_{\ell}^{1}),\cdots,\sum_{i,j}\delta(\mathbf{f}_{1,\ell}^{(i,j)},\mathbf{c}_{\ell}^{D})],
\end{equation}
where $\delta(\mathbf{f}_{1,\ell}^{(i,j)},\mathbf{c}_{\ell}^{d})=1$ if the nearest visual word of $\mathbf{f}_{1,\ell}^{(i,j)}$ is $\mathbf{c}_{\ell}^{d}$, else $\delta(\mathbf{f}_{1,\ell}^{(i,j)},\mathbf{c}_{\ell}^{d})=0$. Finally, the deep pyramid matching kernel for two images $\mathbf{I}_{1}$ and $\mathbf{I}_{2}$ is obtained as follows:
\begin{equation}\label{k}
 K(\mathbf{I}_{1},\mathbf{I}_{2}) = \sum_{\ell=1}^{L}\omega_{\ell}K_{\ell}(\mathbf{I}_{1},\mathbf{I}_{2}),
\end{equation}
where $L$ is the total number of convolutional layers, $\omega_{\ell}$ is the fusion
weight of $\ell-th$ layer,
and $K_{\ell}(\mathbf{I}_{1},\mathbf{I}_{2}) = \sum_{d=1}^{D}\min(\sum_{i,j}\delta(\mathbf{f}_{1,\ell}^{(i,j)},\mathbf{c}_{\ell}^{d}),\sum_{i,j}\delta(\mathbf{f}_{2,\ell}^{(i,j)},\mathbf{c}_{\ell}^{d}))$.

For remote sensing scene classification, we also need to consider the label information for training images. Therefore, instead of using the pre-defined values, we adaptively learn the optimal weights $w_{\ell}$ from training data itself. As discussed in \cite{he2008}, the kernel matrix $\mathbf{K}$ of training data should be close to the ideal matrix $\mathbf{Y}$. The element $\mathbf{K}_{i,j}$ in $\mathbf{K}$ defines the deep pyramid matching kernel for two images $\mathbf{I}_{i}$ and $\mathbf{I}_{j}$, and can be calculated according to Eq.~(\ref{k}). The element $\mathbf{Y}_{i,j}=1$ if the image labels $y_{i}=y_{j}$; otherwise $\mathbf{Y}_{i,j}=0$.
%whose elements $\mathbf{Y}_{i,j}=1$ if the image labels $y_{i}=y_{j}$; otherwise $\mathbf{Y}_{i,j}=0$.
An alternative scheme is to minimize $\|\mathbf{K}-\mathbf{Y}\|_{F}^{2}$, which represents the sum of element-wise distance between matrices $\mathbf{K}$ and $\mathbf{Y}$, i.e., $\|\mathbf{K}-\mathbf{Y}\|_{F}^{2} = \sum_{i}\sum_{j}(\mathbf{K}_{i,j}-\mathbf{Y}_{i,j})^{2}$. Besides, a regularization term $\|\mathbf{w}\|_{2}^{2}$ comprised by all the weights $\omega_{\ell}, \ell\in\{1,\cdots,L\}$ is used to prevent overfitting. Therefore, the objective function of ADPM is formulated as follows:
\begin{gather}\label{objective_function}
\notag
\min_{\mathbf{w}}\,\|\mathbf{K}-\mathbf{Y}\|_{F}^{2}+\lambda\|\mathbf{w}\|_{2}^{2},\\
 s.t. \,\, \mathbf{K} = \sum_{\ell=1}^{L}\omega_{\ell}\mathbf{K}_{\ell}, \,\sum_{\ell=1}^{L}\omega_{\ell}=1,\\
 \notag
 \omega_{\ell}\geq0, \ell = 1,2,\cdots,L.
\end{gather}

Replacing $\mathbf{K}$ by $\sum_{\ell=1}^{L}\omega_{\ell}\mathbf{K}_{\ell}$, we can derive $\|\mathbf{K}-\mathbf{Y}\|_{F}^{2}=\mathbf{w}^{\top}\mathbf{A}\mathbf{w}-2\mathbf{b}^{\top}\mathbf{w}+c$. The element of matrix $\mathbf{A}$ is $\mathbf{A}_{i,j}=\textmd{tr}(\mathbf{K}_{i}^{\top}\mathbf{K}_{j})$, the element of vector $\mathbf{b}$ is $\mathbf{b}_{j}=\textmd{tr}(\mathbf{Y}^{\top}\mathbf{K}_{j})$, and $c=\textmd{tr}(\mathbf{Y}^{\top}\mathbf{Y})$, where `$\textmd{tr}$' denotes the trace of matrices. Thus, Eq.~(3) can be rewritten as a canonical quadratic programming problem:
\begin{gather}\label{final}
\notag
\min_{\mathbf{w}}\,\mathbf{w}^{\top}(\mathbf{A}+\lambda \mathbf{I})\mathbf{w}-2\mathbf{b}^{\top}\mathbf{w},\\
 s.t. \,\, \omega_{\ell}\geq0, \ell = 1,2,\cdots,L,\\
 \notag
 \sum_{\ell=1}^{L}\omega_{\ell}=1.
\end{gather}

We can use the \textit{quadprog} function in MATLAB to directly optimize Eq.~(4).
Subsequently, the optimal weight $\mathbf{w}$ is employed to calculate the kernel matrix $\mathbf{K}$, which is then fed to the SVM for classification.
Finally, the classification results derived from all scales are integrated via the majority voting method to achieve the final classification results.

\section{Experimental Results}

To evaluate the effectiveness of the proposed method, we compare it with several state-of-the-art methods on two widely used remotely sensed scene datasets: \textit{21-Class-Land-Use} dataset \cite{yang2011} and \textit{19-Class Satellite Scene} dataset \cite{xia2010,dai2011}.

\subsection{21-Class-Land-Use dataset}
\subsubsection{\textbf{Dataset description and experimental setup}}
\begin{figure*}
  \centering
  \includegraphics[scale=0.7]{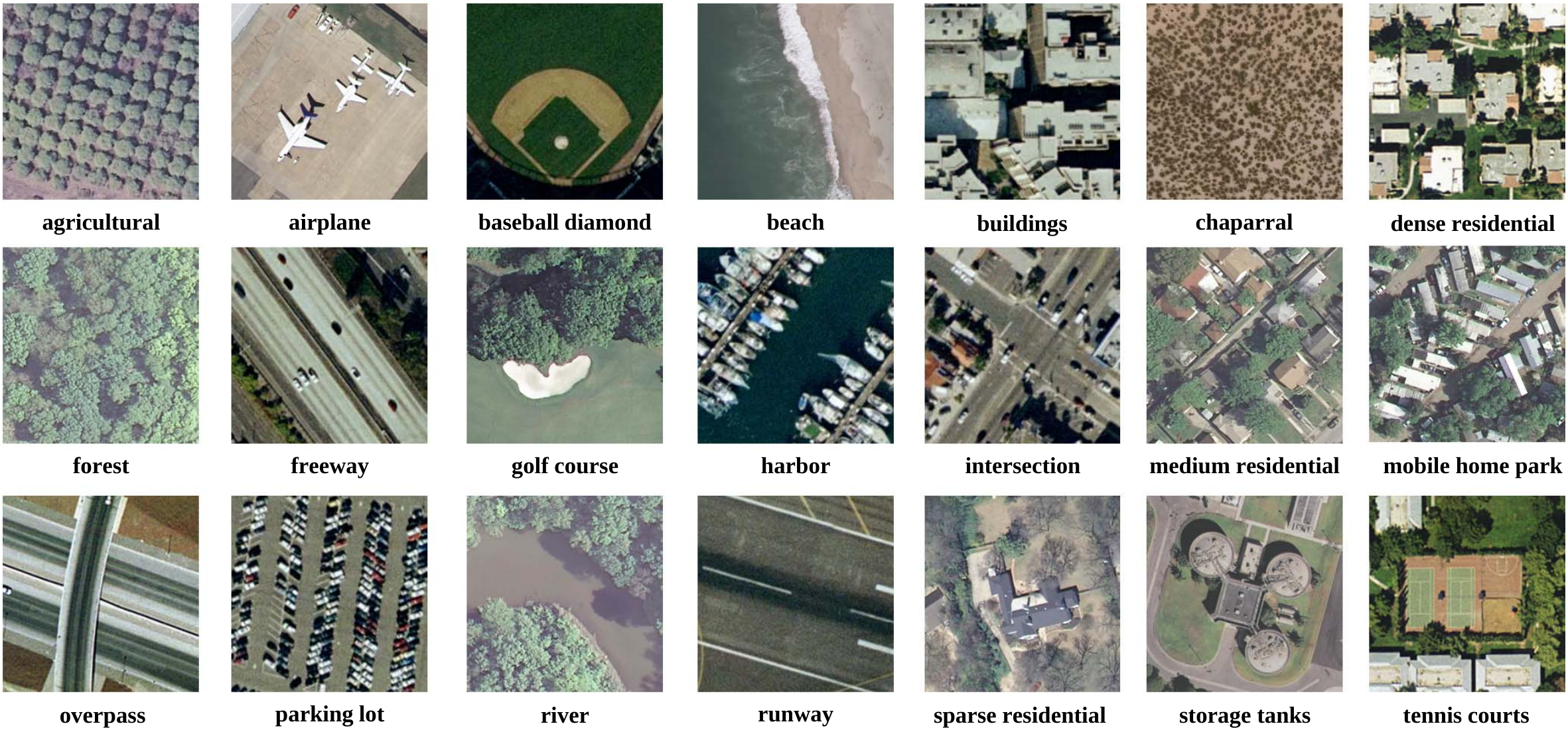}\\
  \caption{An image example from each class in 21-Class-Land-Use dataset.}\label{21classes}
\end{figure*}

This dataset was manually extracted from aerial orthoimagery downloaded from the United States Geological Survey (USGS) National Map. It consists of 21 different land use and land cover classes, including \textit{agricultural, airplane, baseball diamond, beach, buildings, chaparral, dense residential, forest,
freeway, golf course, harbor, intersection, medium density residential, mobile home park, overpass, parking lot, river, runway, sparse residential, storage tanks} and \textit{tennis courts}. Each class contains 100 RGB images with spatial resolution of one foot (i.e., 0.3 m) and image size of $256\times256$ pixels. Fig.~\ref{21classes} demonstrates an image example from each class.

In each experiment, besides the original scale, the images are warped into three different scales, including $128\times128$, $192\times192$ and $227\times227$ pixels.
%We choose $227\times227$ pixels because the training of SPP-net needs to take advantage of the pre-training results from CaffeNet \cite{jia2014}, which requires this input scale.
We choose $227\times227$ pixels to be consistent with the input scale of the pre-training CaffeNet \cite{jia2014}.
The dataset is randomly divided into two sets: the training set and the testing set. The training set is used to fine-tune the fully-connected layers in the SPP-net. For the pixels in each convolutional layer, $k$-means clustering is employed to form the visual codebook with 300 code words. The encoded deep features by ADPM are then fed into SVM classifiers with histogram intersection kernels, which are implemented using the LIBSVM package \cite{Chang2011}, and one-against-one strategy is adopted to address the multi-class issue. The testing set is used to evaluate the performance of classifiers. In order to reduce the effect of random selection, we repeat each algorithm execution on ten different training/testing split of the data set and report means and standard deviations of the obtained accuracies. The parameter $\lambda$ in Eq.~(3) is set to 0.5 empirically.

\subsubsection{\textbf{Convolutional vs. Fully-connected layers}}
\begin{figure}
  \centering
  \includegraphics[scale=0.5]{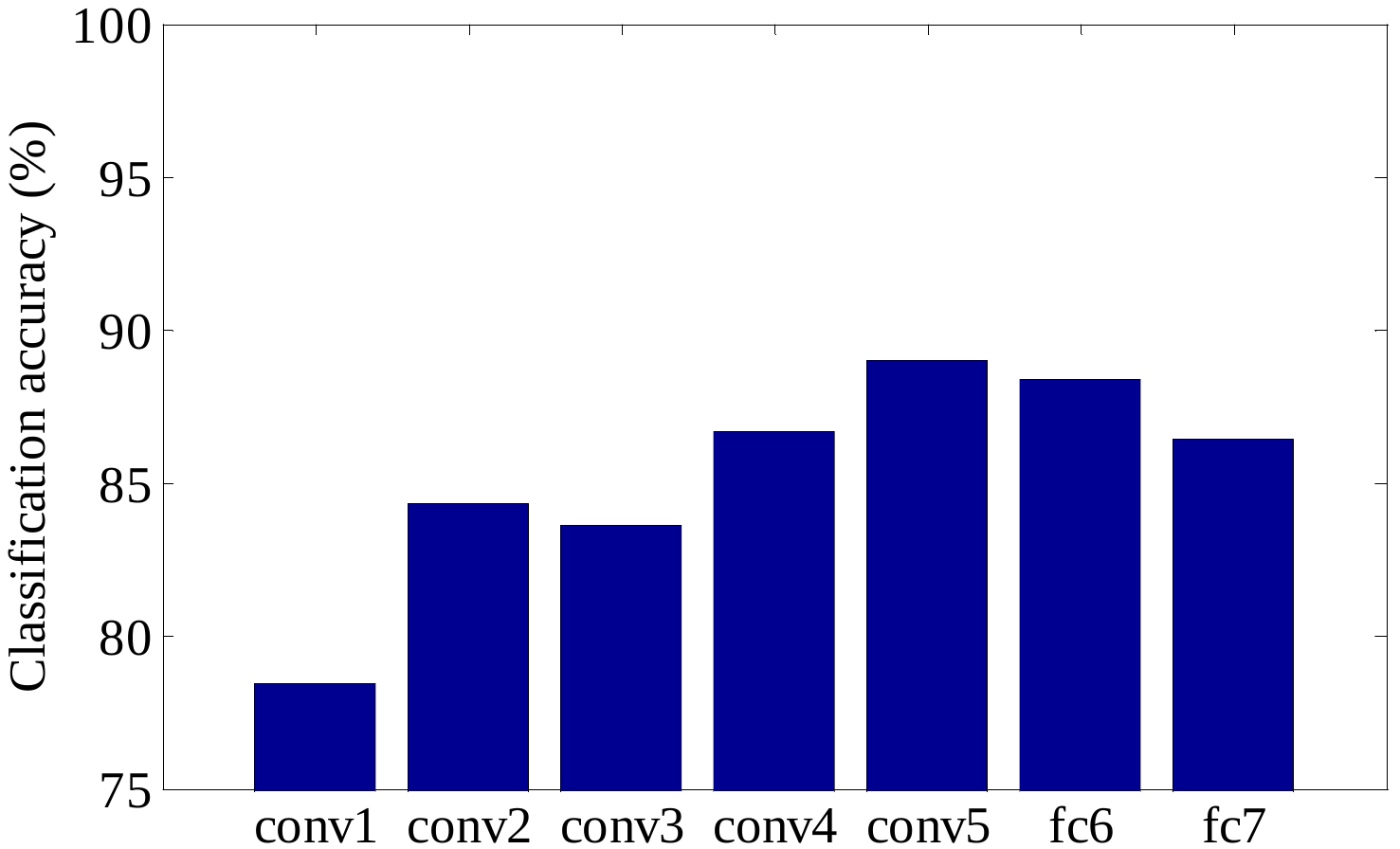}\\
  \caption{Performance of each layer features in SPP-net on 21-Class-Land-Use dataset.}\label{21-EachLayer}
\end{figure}

To assess the performance of each layer in the SPP-net, we randomly choose $50\%$ samples from each class as the training set, and warp all of the images to $227\times227$ pixels. For simplicity, convolutional and fully-connected layers are abbreviated as \textit{conv} and \textit{fc}, respectively. For instance, conv5 represents the fifth convolutional-layer; fc6 denotes the sixth fully-connected layer. Fig.~\ref{21-EachLayer} demonstrates the classification accuracies of each layer. For the convolutional-layer features, conv5 obtains the best result because deeper convolutional-layers can learn much richer discriminant features. For the fully-connected layers, fc6 gets a higher accuracy than fc7, which is consistent with the conclusion in \cite{girshick2014}. More importantly, conv5 achieves superior result as compared to fc6. This can be explained by two reasons. The first one is that conv5 contains rich spatial information, thus improving the classification performance. The other one is that the fully-connected layers are not well fine-tuned by such a small number of training samples, thus degrading the classification performance.
\begin{figure}
  \centering
  \includegraphics[scale=0.5]{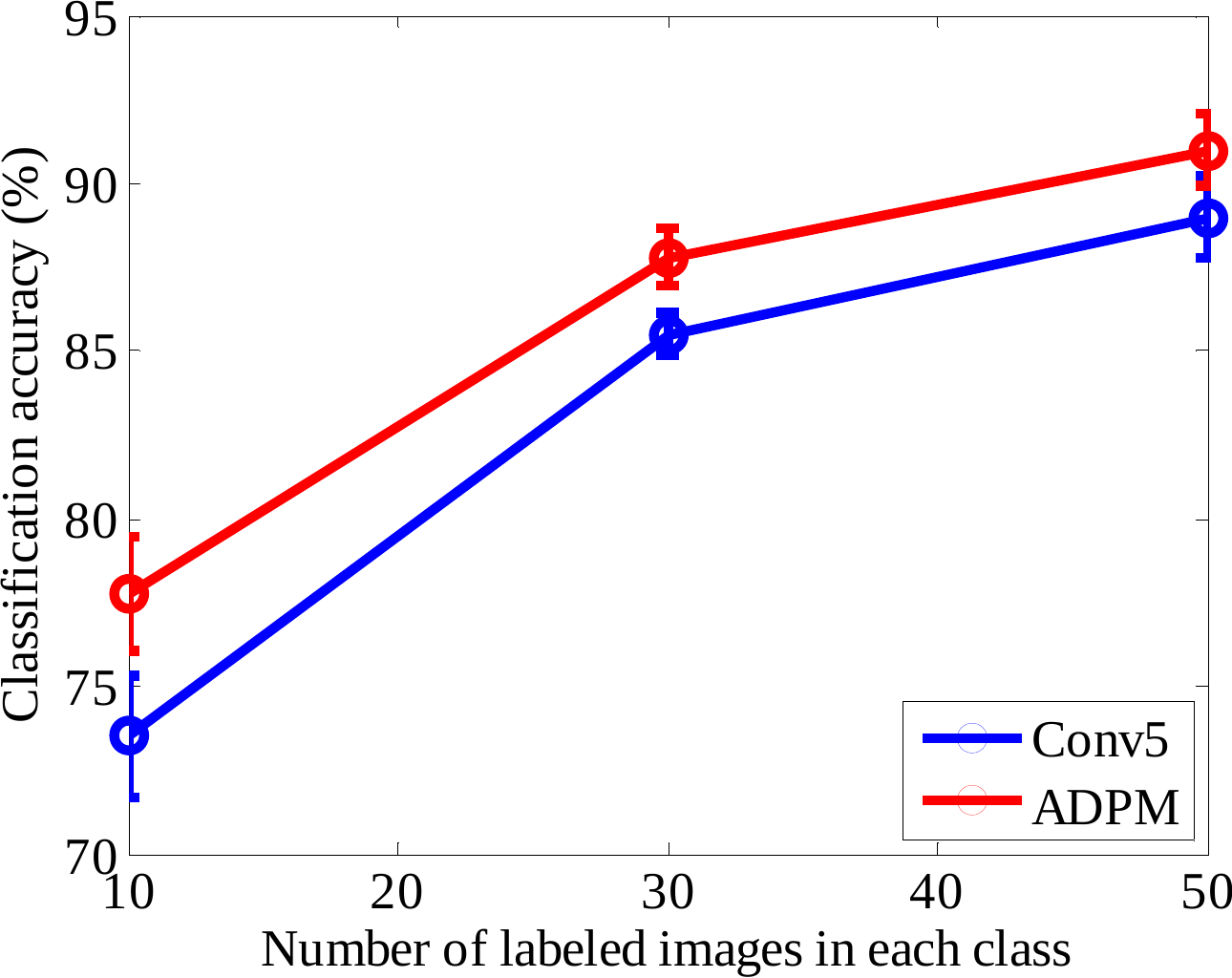}\\
  \caption{Performance comparison using features from conv5 and the proposed ADPM method with different numbers of training samples on 21-Class-Land-Use dataset.}\label{21-conv5-ADPM}
\end{figure}

\begin{figure}
  \centering
  \includegraphics[scale=0.5]{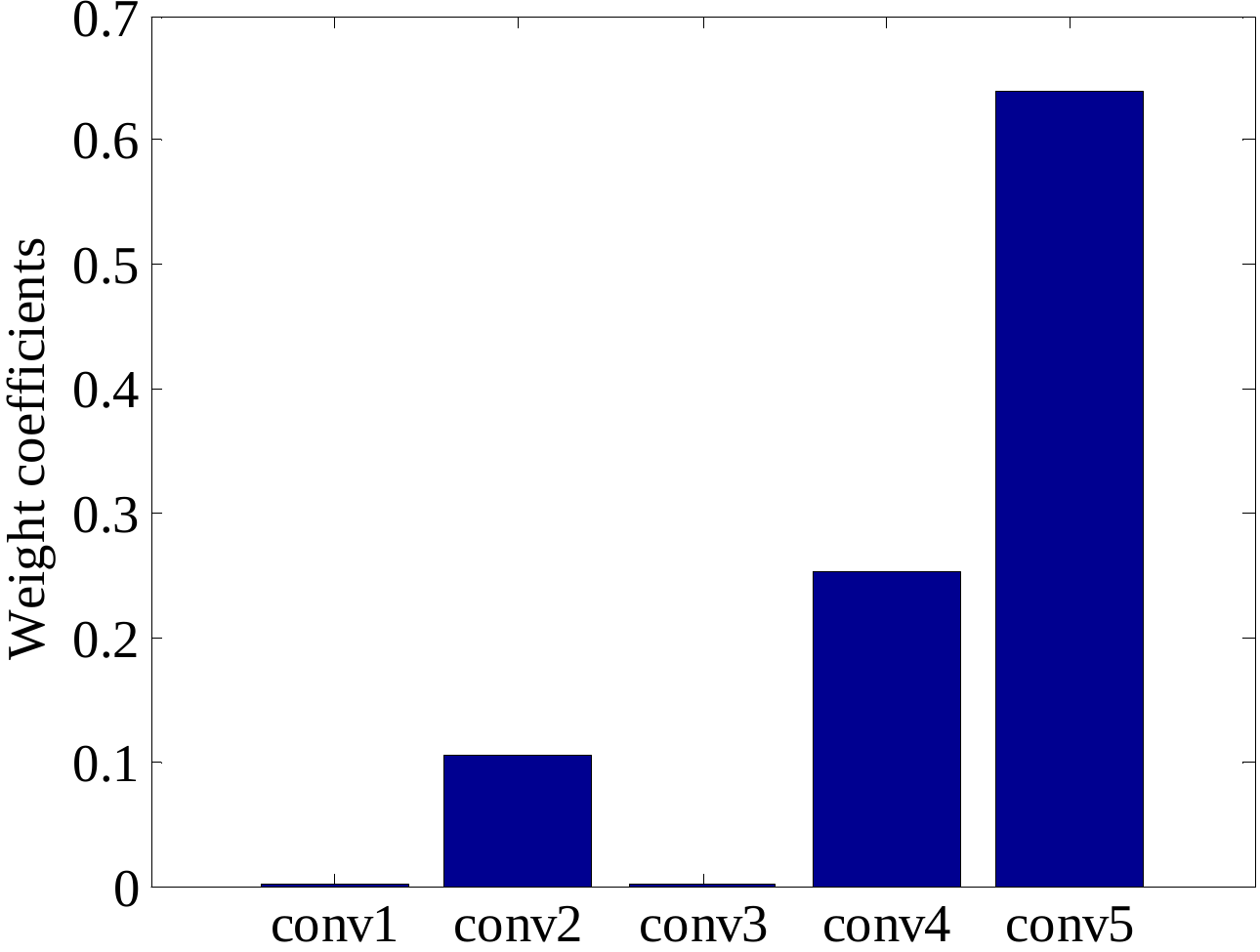}\\
  \caption{Weights of each convolutional layer in ADPM on 21-Class-Land-Use dataset.}\label{21-weights}
\end{figure}

\begin{figure}
  \centering
  \includegraphics[scale=0.5]{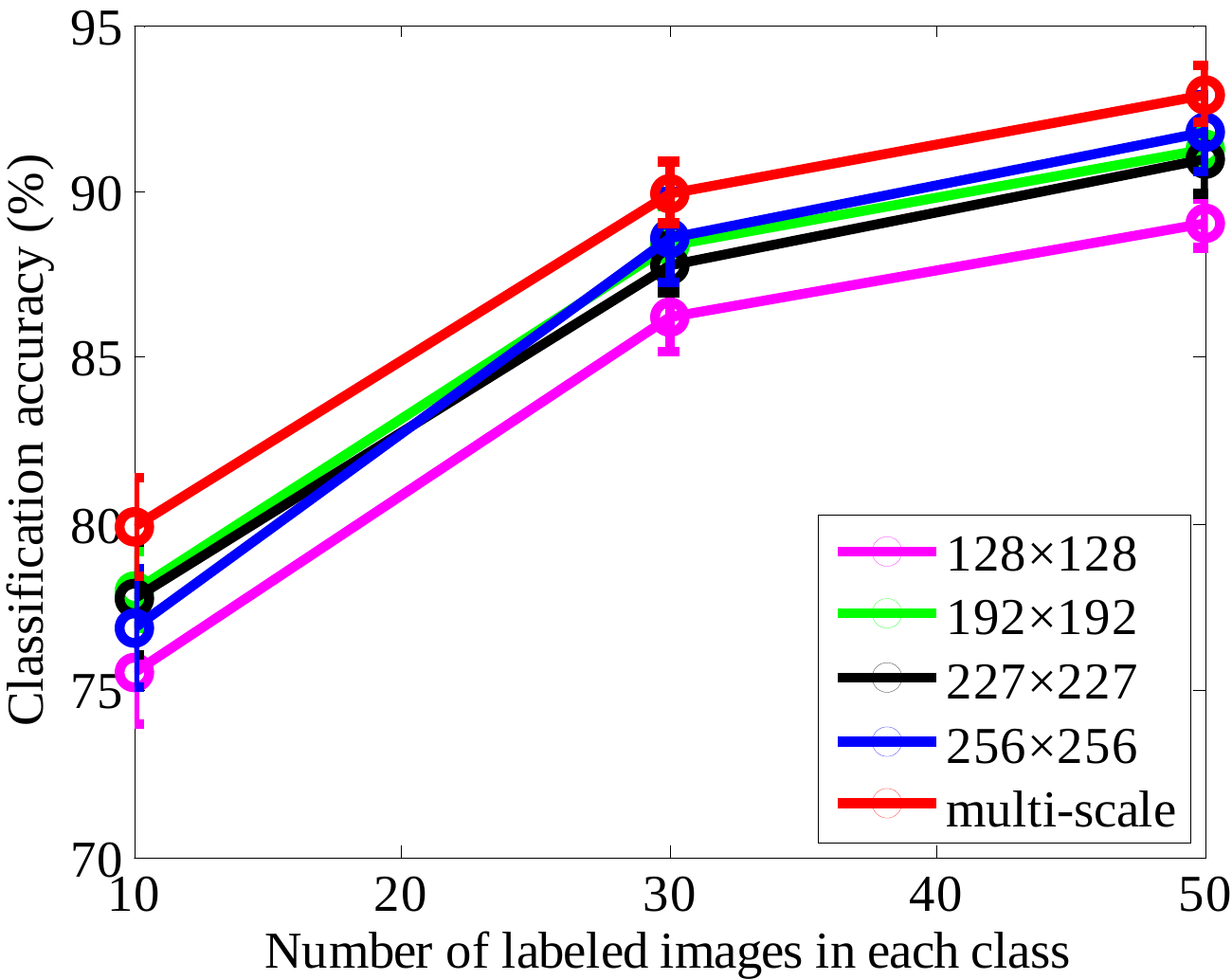}\\
  \caption{Performance comparison between single-scale ADPM and multi-scale ADPM methods with different numbers of training samples on 21-Class-Land-Use dataset.}\label{21-MultiScale}
\end{figure}

To validate the effectiveness of the proposed ADPM method, we compare it with conv5 using different numbers of training samples with size of $227\times227$ pixels. Fig.~\ref{21-conv5-ADPM} demonstrates the detailed comparison results. It is shown that as the number of the training samples increases, the performance of ADPM and conv5 improves. Besides, ADPM achieves the superior performance than conv5 with the same number of the training samples. This indicates that different convolutional layers capture different features of the input image, and sufficiently fusing these informations can generate a more powerful representation. Fig.~\ref{21-weights} shows the fusing weights of each convolutional-layer in ADPM. It can be observed that the weight of conv1 is close to 0, because it extracts the edge features of input images, which have less discriminative information. Besides, conv3 and conv4 correspond to representations of input images in the same spatial resolution, while the information in conv4 is more abstract than in conv3. This is the reason why the weight of conv3 is also close to 0. Additionally, we also evaluate the effects of input image scales on performance, which is demonstrated in Fig.~\ref{21-MultiScale}. When the number of training samples is 10, the best scale is $192\times192$ pixels. When it exceeds 30, the best scale is $256\times256$ pixels. However, the multi-scale ADPM always achieves the highest accuracies, because it can integrate the complementary information from different scales.

\subsubsection{\textbf{Comparison with state-of-the-art methods}}
\begin{figure}
  \centering
  \includegraphics[scale = 0.45]{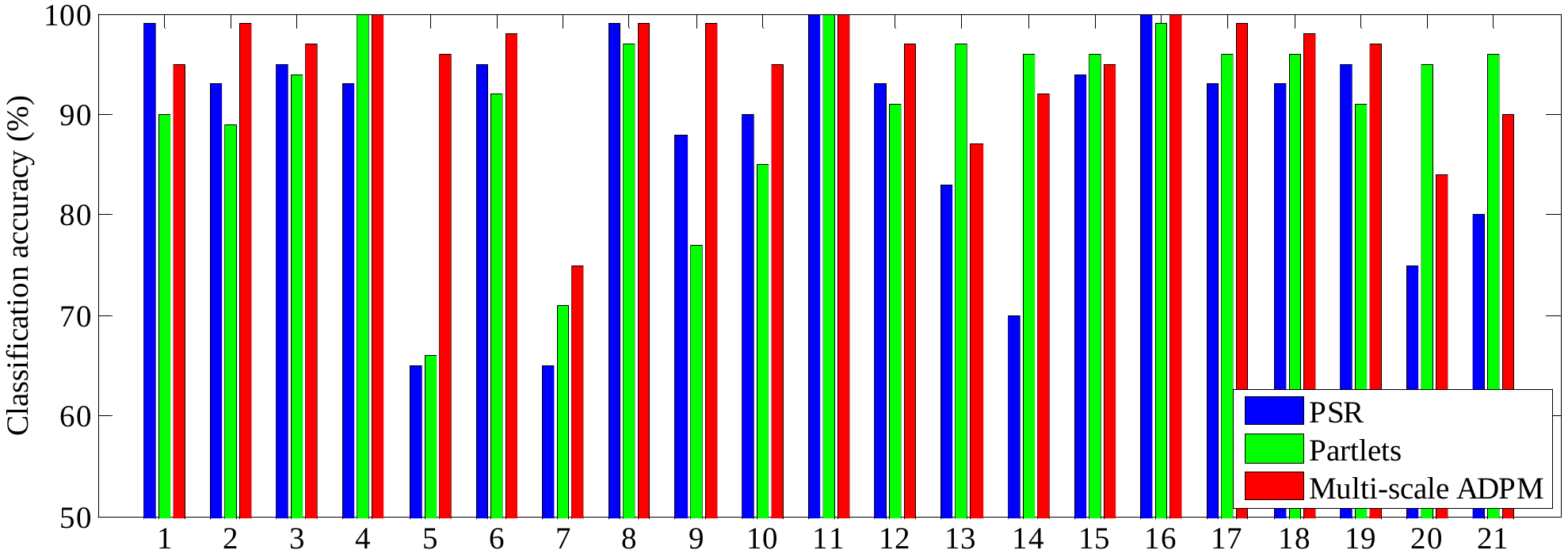}\\
  \caption{Each class accuracy of multi-scale ADPM and two state-of-the-art methods. The class labels are assigned as follows: 1 = Agricultural, 2 = Airplane, 3 = Baseball diamond, 4 = Beach, 5 = Buildings, 6 = Chaparral, 7 = Dense residential, 8= Forest, 9 = Freeway, 10 = Golf course, 11 = Harbor, 12 = Intersection, 13 = Medium residential, 14 =Mobile home park, 15 = Overpass, 16 = Parking lot, 17 = River, 18 = Runway, 19 = Sparse residential, 20= Storage tanks, and 21 = Tennis courts.}\label{21-EachClassAcc}
\end{figure}

To demonstrate the superiority of the proposed method, we compare it with several state-of-the-art approaches, including SPMK \cite{lazebnik2006}, BoVW \cite{yang2011}, SPCK++ \cite{yang2011}, BRSP \cite{jiang2012}, SC+Pooling \cite{cheriyadat2014}, UFL \cite{hu2015UFL}, Partlets-based method \cite{cheng2015} and PSR \cite{chen2015}. All the experiments are conducted using 5-fold cross validation. The classification accuracies are shown in Table~\ref{21-state-of-the-art-comparison}, where the results of the state-of-the-art methods are the reported results in relevant references.
%Table~\ref{21-state-of-the-art-comparison} reports the classification accuracies of all state-of-the-art methods, as they appear in the original papers, together with the accuracies of our proposed methods.
From this table, we can observe that the three single-scale ADPM methods are better than the state-of-the-art methods, and the multi-scale ADPM can further improve the performance by fusing the classification results from them. Specifically, the proposed multi-scale ADPM guarantees a large performance gain as compared to all the state-of-the-art methods, with a minimum gap of almost 4\%. Besides, Fig.~\ref{21-EachClassAcc} shows each class accuracies of the multi-scale ADPM and two state-of-the-art methods: PSR and Partlets. Multi-scale ADPM achieves the highest accuracies in 15 classes. The most difficult class is \textit{dense residential}, whose accuracy is less than 80\%, because it has very similar structures with \textit{medium residential} and \textit{mobile home park} classes.

\begin{table}\small
\centering
\caption{Classification accuracies of state-of-the-art methods and the proposed methods on 21-Class-Land-Use dataset.}
\label{21-state-of-the-art-comparison}
\begin{tabular}{|c|c|c|c|}
\hline
\multicolumn{4}{|c|}{\textbf{State-of-the-art Methods}}    \\ \hline
Method   & Accuracy & Method           & Accuracy \\ \hline
SPMK \cite{lazebnik2006}    & 74.00    & BoVW \cite{yang2011}            & 71.68    \\ \hline
SPCK++ \cite{yang2011}  & 77.38    & BRSP \cite{jiang2012}            & 77.80    \\ \hline
UFL  \cite{hu2015UFL}    & 90.26    & SC+Pooling \cite{cheriyadat2014}      & 81.67    \\ \hline
Partlets \cite{cheng2015} & 91.33    & PSR  \cite{chen2015}            & 89.10    \\ \hline
\multicolumn{4}{|c|}{\textbf{Proposed Methods}}   \\ \hline
Method   & Accuracy & Method           & Accuracy \\ \hline
ADPM-192 & 92.67    & ADPM-227         & 92.04    \\ \hline
ADPM-256 & 93.52    & Multi-scale ADPM & 94.86    \\ \hline
\end{tabular}
\end{table}

\subsection{19-Class Satellite Scene dataset}
\subsubsection{\textbf{Dataset description and experimental setup}}
\begin{figure*}
  \centering
  \includegraphics[scale=0.7]{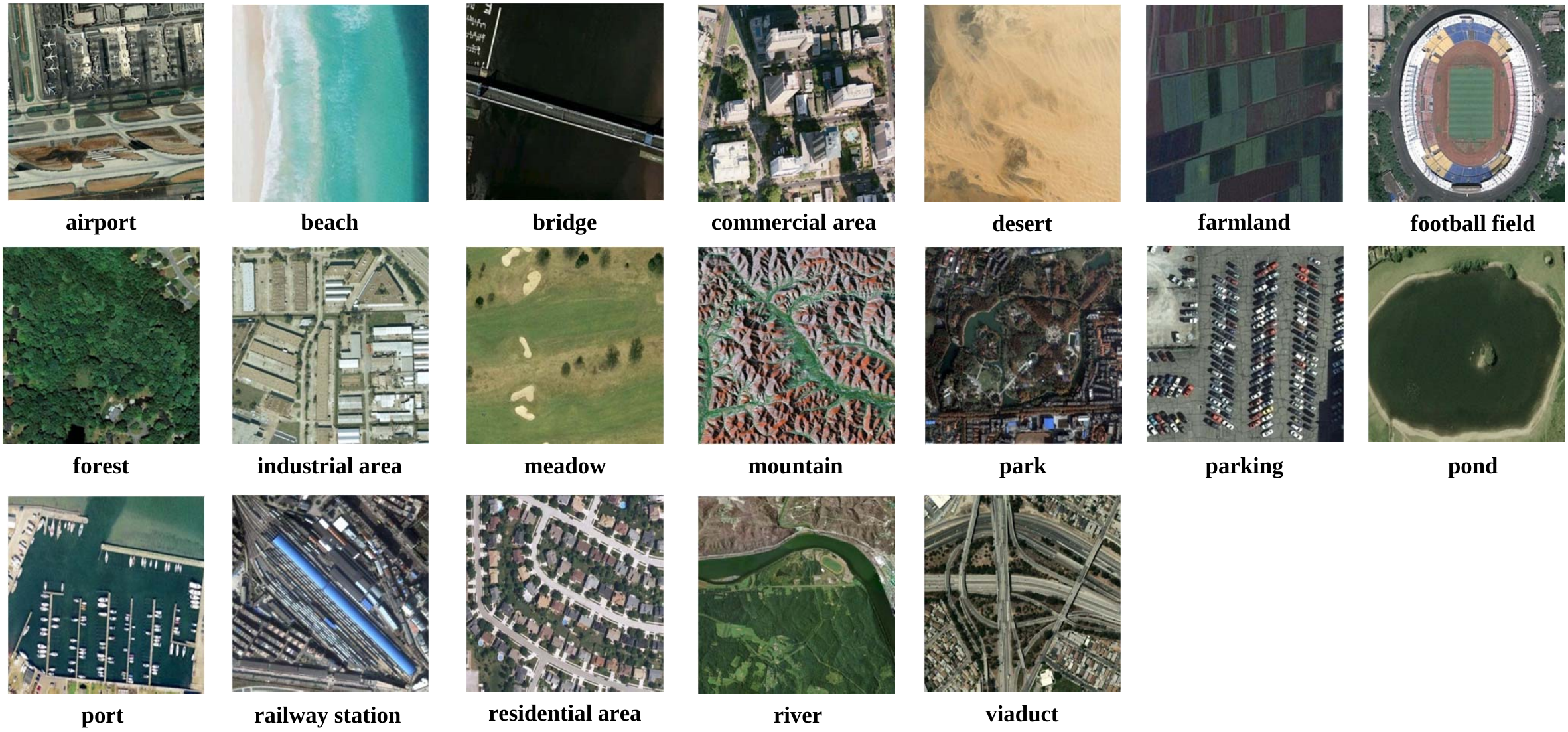}\\
  \caption{An image example from each class in 19-Class Satellite Scene dataset.}\label{19classes}
\end{figure*}

The second dataset is composed of 19 classes of scenes, including \textit{airport, beach, bridge, commercial area, desert, farmland, football field, forest, industrial area, meadow, mountain, park, parking, pond, port, railway station, residential area, river} and \textit{viaduct}. Each class has 50 images, with size of $600\times600$ pixels. These images are extracted from very large satellite images on Google Earth. Fig.~\ref{19classes} shows an image example from each class. The dataset is randomly divided into two sets: the training set and the testing set. Similar to 21-Class-Land-Use dataset, the original images are warped into different scales: $128\times128$, $192\times192$, $227\times227$, $256\times256$ and $384\times384$ pixels. The visual codebook in BoVW is generated by $k$-means clustering with 300 centers. The parameter $\lambda$ in Eq. (3) is set to 0.5 empirically. In classification phase, the encoded convolutional-layer features are fed into SVM classifiers with the histogram intersection kernels.  To obtain stable results, all experiments are repeated 10 times with different training/testing split. The final performance is reported as the mean and standard deviation of the results from 10 runs.

\subsubsection{\textbf{Convolutional vs. Fully-connected layers}}
\begin{figure}
  \centering
  \includegraphics[scale=0.5]{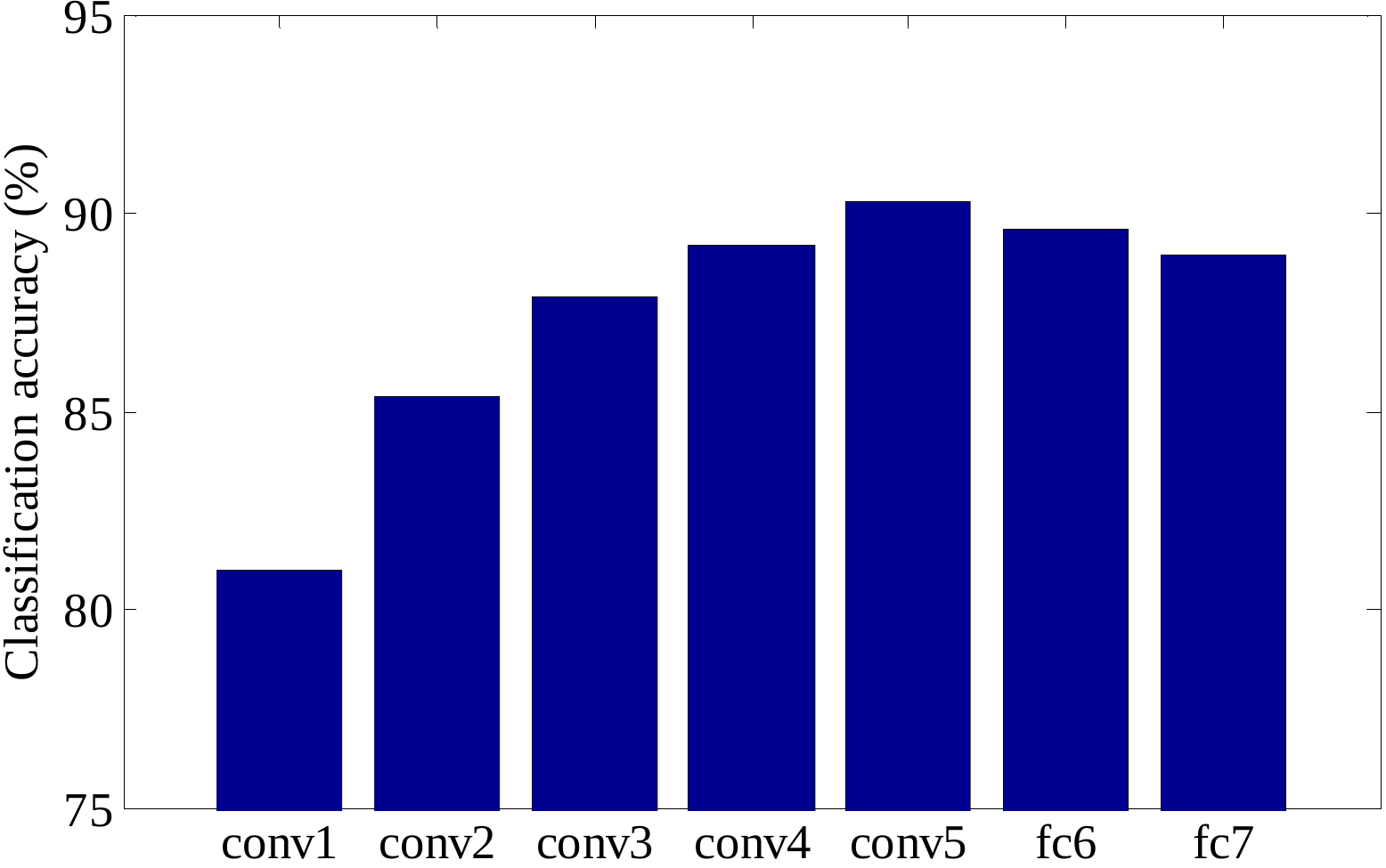}\\
  \caption{Performance of each layer features in SPP-net on 19-Class Satellite Scene dataset.}\label{19-EachLayer}
\end{figure}

\begin{figure}
  \centering
  \includegraphics[scale=0.5]{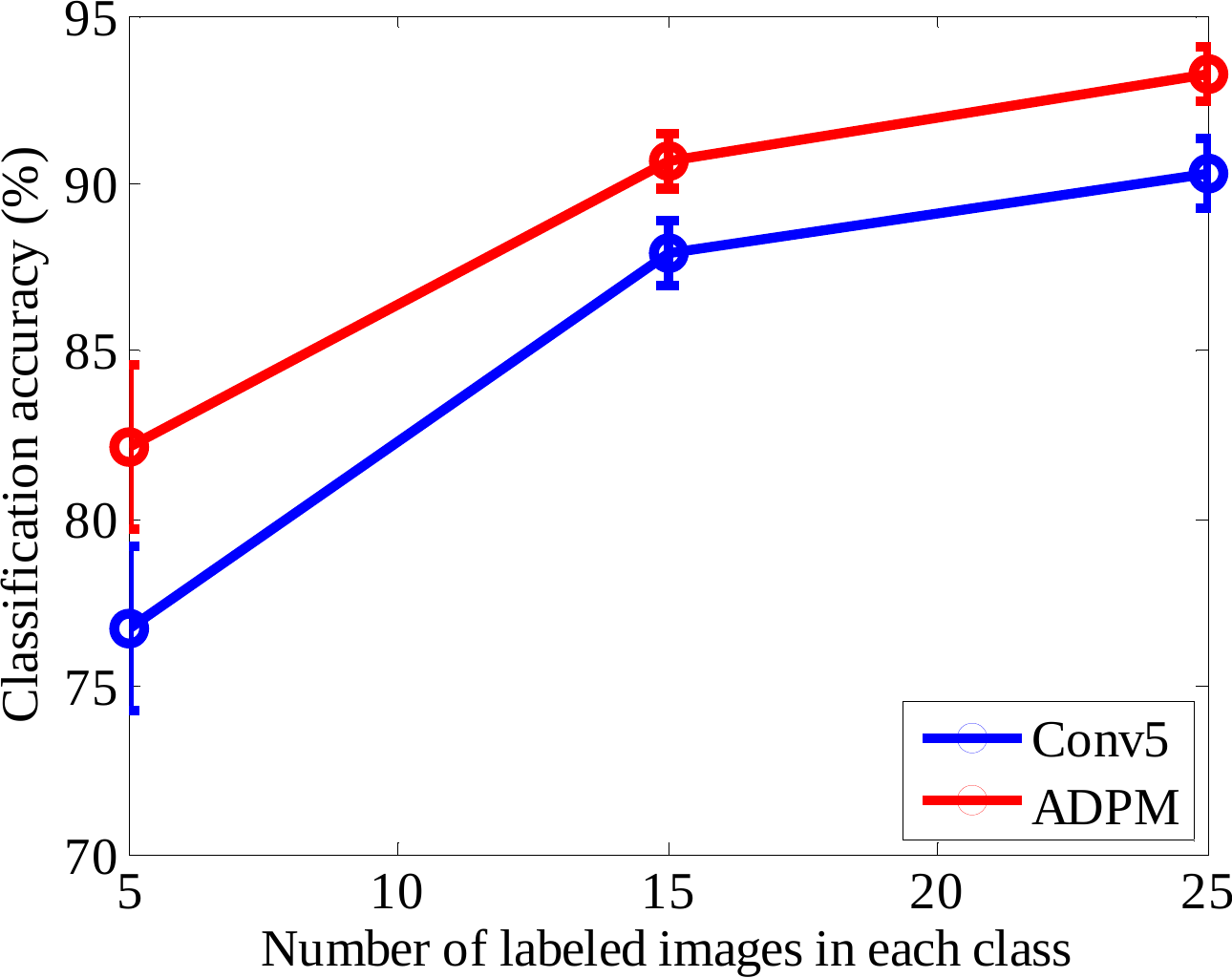}\\
  \caption{Performance comparison using features from conv5 and the proposed ADPM method with different numbers of training samples on 19-Class Satellite Scene dataset.}\label{19-conv5-ADPM}
\end{figure}

\begin{figure}
  \centering
  \includegraphics[scale=0.5]{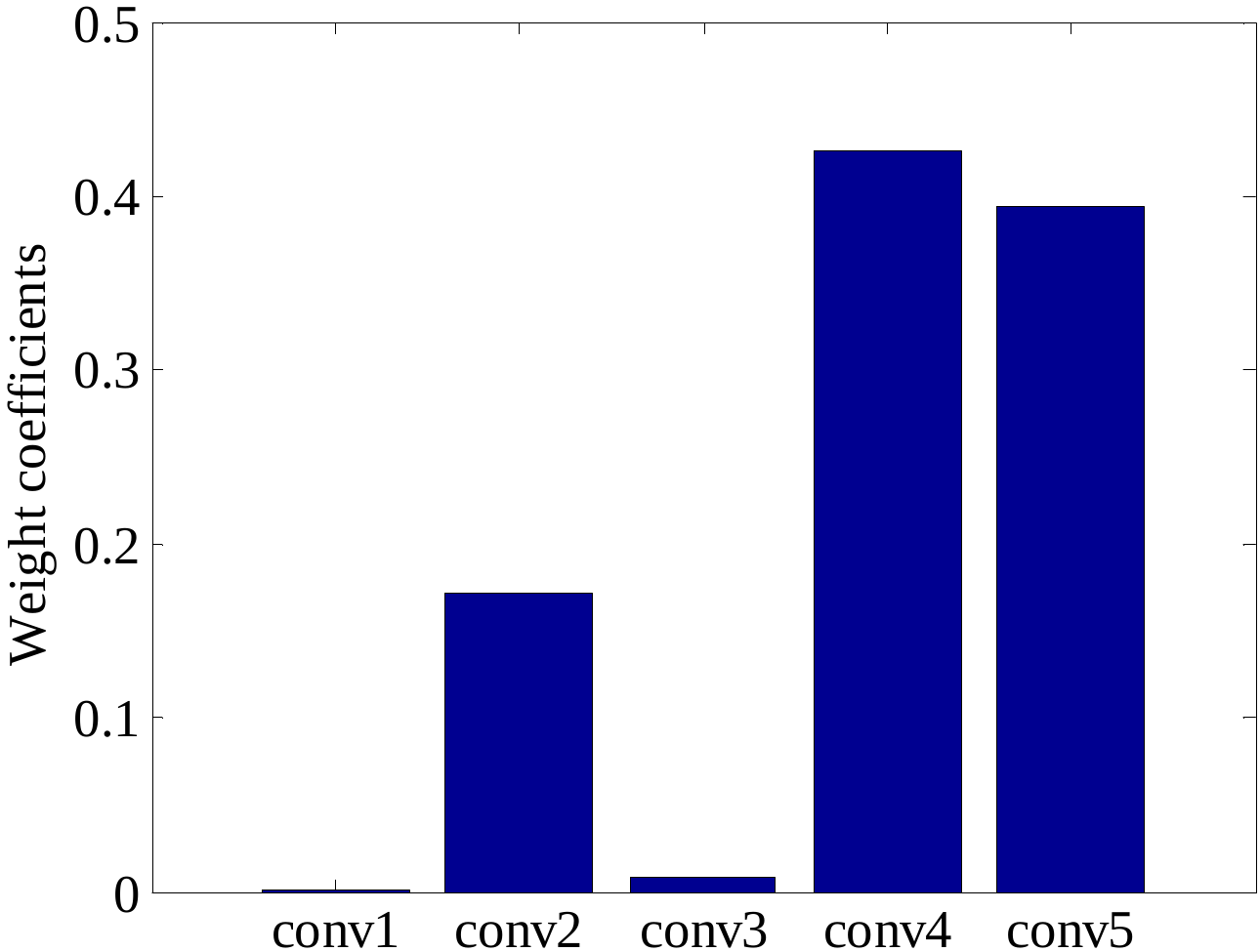}\\
  \caption{Weights of each convolutional layer in ADPM on 19-Class Satellite Scene dataset.}\label{19-weights}
\end{figure}

\begin{figure}
  \centering
  \includegraphics[scale=0.5]{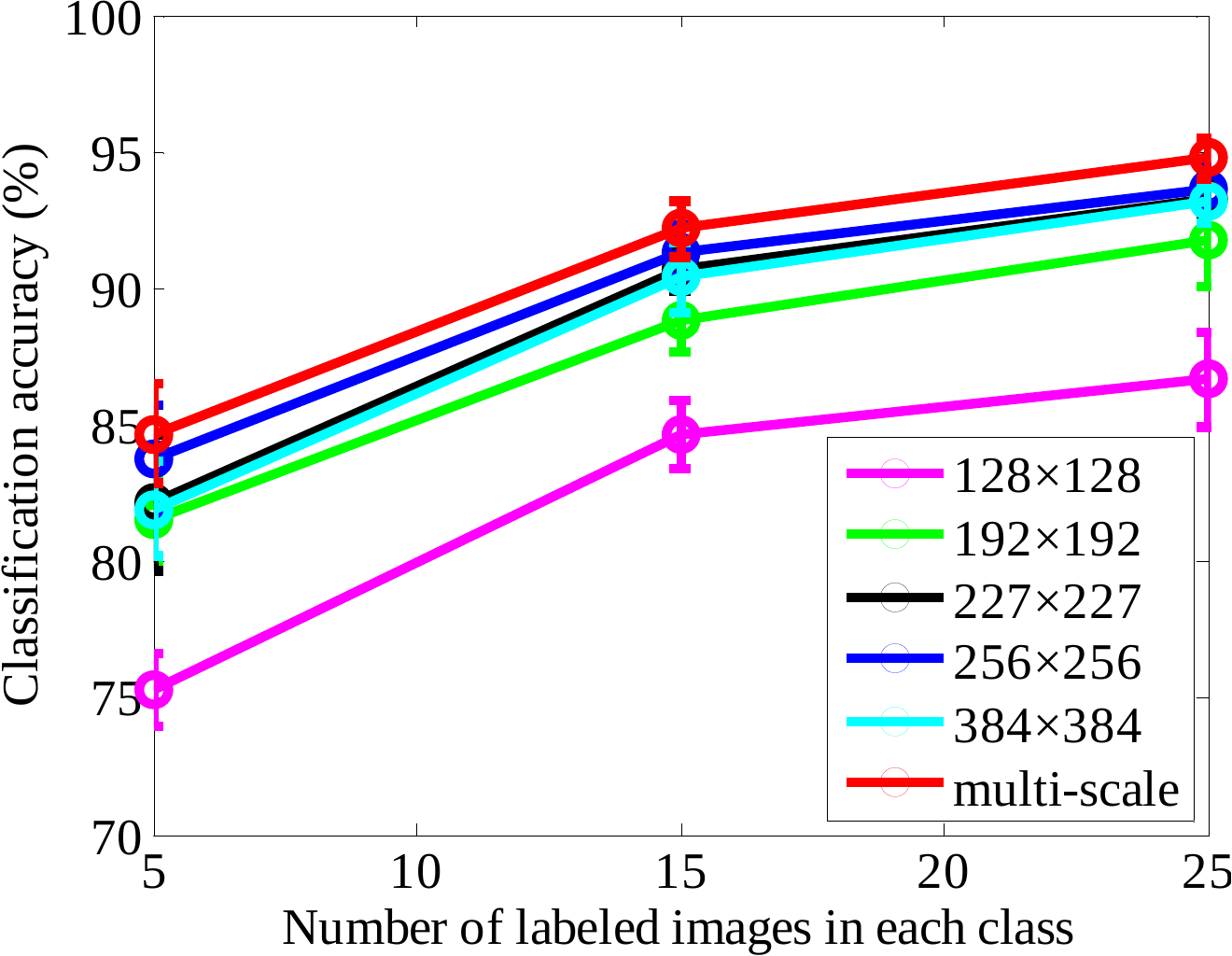}\\
  \caption{Performance comparison between single-scale ADPM and multi-scale ADPM methods with different numbers of training samples on 19-Class Satellite Scene dataset.}\label{19-MultiScale}
\end{figure}

Fig. \ref{19-EachLayer} demonstrates the classification accuracies of each layer features in the SPP-net using 50\% samples from each class and $227\times227$ input pixels. Same as the results on 21-Class-Land-Use dataset, conv5 achieves superior performance compared to fully-connected layers. To validate the effectiveness of the proposed method, we compare the performance of ADPM and conv5 using different numbers of training samples. As shown in Fig.~\ref{19-conv5-ADPM}, with the same number of training samples, ADPM is significantly better than conv5 in terms of classification accuracy. Fig.~\ref{19-weights} analyzes the contribution of each convolutional-layer in ADPM. We can observe that the weight of conv1 is close to 0, because it aims to extract the general features like edges, which captures a few discriminative information. Besides, conv3 has a small contribution to ADPM. It is caused by the fact that conv3 and conv4 extract the same resolution features, while conv4 contains most of the discriminative information from conv3. In addition,  Fig.~\ref{19-MultiScale} shows the performance comparison between single-scale ADPM and multi-scale ADPM methods under different numbers of training samples. It is observed that $256\times256$ input pixels obtain the best results for single-scale ADPM methods, and multi-scale ADPM can further improve the performance by fusing the complementary information from different scales. It is worth noting that we only use $227\times227$, $256\times256$ and $384\times384$ input pixels to construct the multi-scale ADPM, because the accuracies achieved by $128\times128$ and $192\times192$ pixels are low, and directly combining all of these results degrades the final performance.

\subsubsection{\textbf{Comparison with state-of-the-art methods}}
\begin{table}\small
\centering
\caption{Classification accuracies of state-of-the-art methods and the proposed methods on 19-Class Satellite Scene dataset.}
\label{19-state-of-the-art-comparison}
\begin{tabular}{|c|c|}
\hline
Method           & Accuracy \\ \hline
SCMF \cite{sheng2012}            & 78.32    \\ \hline
MKL  \cite{cusano2014}            & 67.32    \\ \hline
SSEP \cite{yang2015}            & 73.82    \\ \hline
ADPM-227         & 82.14    \\ \hline
ADPM-256         & 83.71    \\ \hline
ADPM-384         & 81.91    \\ \hline
Multi-scale ADPM & 84.67    \\ \hline
\end{tabular}
\end{table}
\begin{figure}
  \centering
  \includegraphics[scale = 0.55]{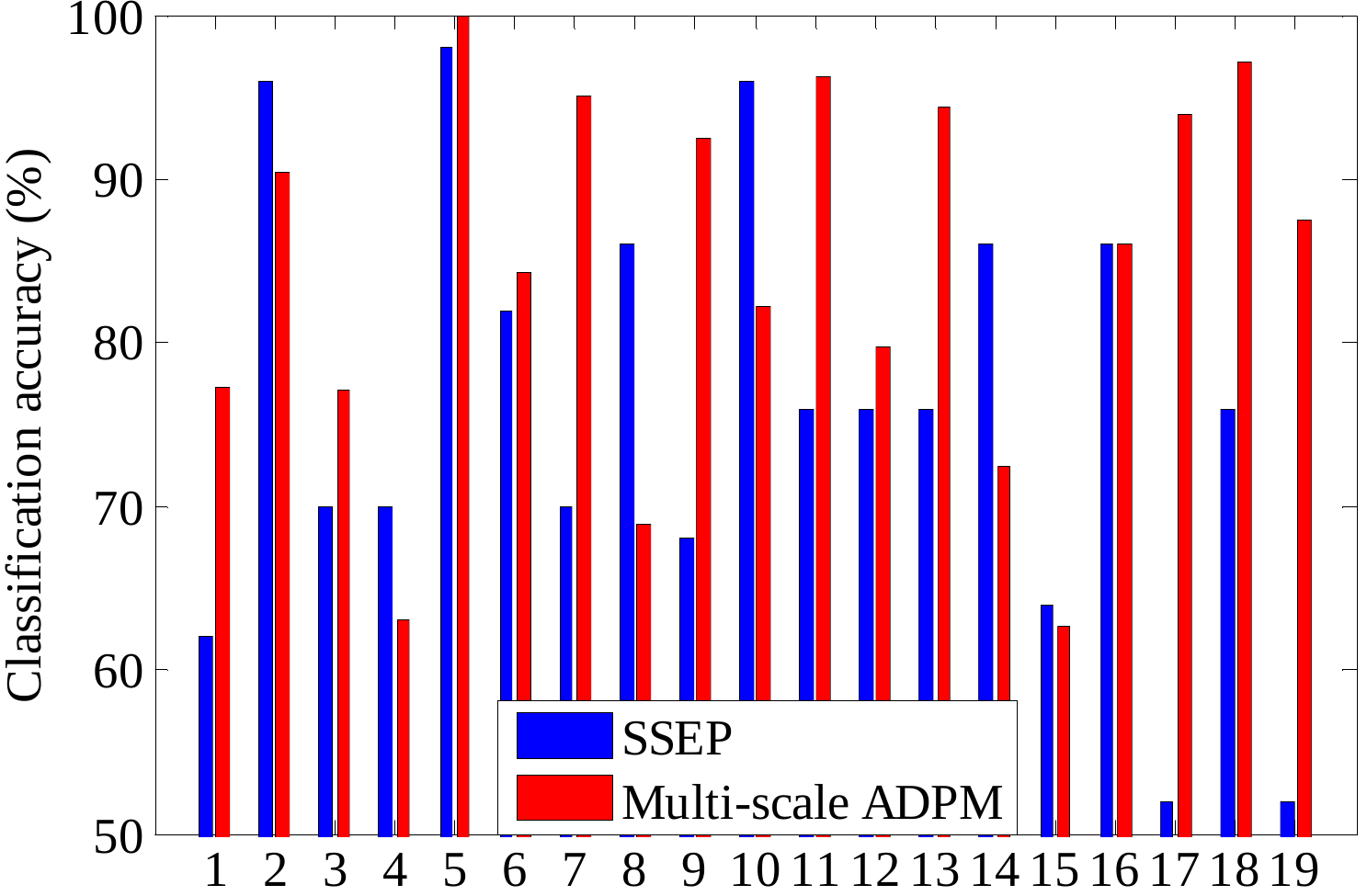}\\
  \caption{Each class accuracies of multi-scale ADPM and SSEP. The class labels are assigned as follows: 1 = Airport, 2 = Beach, 3 = Bridge, 4 = Commercial area, 5 = Desert, 6 = Farmland, 7 = Football field, 8 = Forest, 9 = Industrial area, 10 = Meadow, 11 = Mountain, 12 = Park, 13 = Parking, 14 =  Pond, 15 = Port, 16 = Railway station, 17 = Residential area, 18 = River, 19 = Viaduct.}\label{19-EachClassAcc}
\end{figure}

In order to comprehensively analyze the superiority of the proposed method, we compare it with three state-of-the-art approaches ever tested on this dataset, including SCMF \cite{sheng2012}, MKL \cite{cusano2014} and SSEP \cite{yang2015}. Table~\ref{19-state-of-the-art-comparison} reports the classification accuracies achieved by different methods with 5 training samples from each class. All of the single-scale ADPM models are better than the best state-of-the-art method SCMF, and the multi-scale ADPM can further increase the accuracy as compared to the single-scale models. In particular, the multi-scale ADPM guarantees a large performance gain as compared to all the state-of-the-art methods, with a minimum gap of almost 8\%. Besides, Fig.~\ref{19-EachClassAcc} demonstrates each class accuracies of multi-scale ADPM and SSEP, which indicates that the multi-scale ADPM gets better results in most of classes especially in \textit{residential area} and \textit{viaduct}.

\section{Conclusions}

This paper introduces a new adaptive deep pyramid matching (ADPM) model intended to properly fuse convolutional-layer features in classification of remotely sensed images using convolutional neural networks (CNNs). Actually, deep descriptors from different convolutional layers can be regarded as a multi-resolution representation of the input image. Motivated by a previously developed spatial pyramid matching (SPM) model, we employ pyramid matching kernel to combine them effectively. However, as opposed to the SPM, the fusing weights among different layers are adaptively learned from data itself, instead of the given values. Besides, using a fixed input image size inevitably results in the loss of discriminative information. To address this important issue, we propose to integrate the complementary information from input images with different sizes. In the classification phase, each scale generates a result, and a majority voting strategy is used to combine multiple results. Our experiments are conducted using two widely used remote sensing image datasets to certify the effectiveness of the proposed method. The experimental results indicate that effectively fusing different convolutional-layer features can generate a more powerful representation, and that multi-scale input images can provide much more discriminative information as compared to single-scale ones.

\bibliography{IEEEfull,DeepSPM}
\bibliographystyle{IEEEbib}
\end{document}